\documentclass[conference]{IEEEtran}
\IEEEoverridecommandlockouts
\usepackage{cite}
\usepackage{amsmath,amssymb,amsfonts}
\usepackage{algorithmic}
\usepackage{graphicx}
\usepackage{textcomp}
\usepackage{xcolor}
\def\BibTeX{{\rm B\kern-.05em{\sc i\kern-.025em b}\kern-.08em
    T\kern-.1667em\lower.7ex\hbox{E}\kern-.125emX}}
\begin{document}

\title{Automated segmentation on the entire cardiac cycle using a deep learning work-flow}

\author{\IEEEauthorblockN{Nicol\'o Savioli \IEEEauthorrefmark{1} Miguel Silva Vieira\IEEEauthorrefmark{1},
Pablo Lamata\IEEEauthorrefmark{1},
Giovanni Montana\IEEEauthorrefmark{1} \IEEEauthorrefmark{2}}
\IEEEauthorblockA{\IEEEauthorrefmark{1}King's College London, Division of Imaging Sciences and Biomedical Engineering, UK}
\IEEEauthorblockA{ \IEEEauthorrefmark{2} WMG, University of Warwick, UK}\\
\texttt{\IEEEauthorrefmark{1}\{nicolo.l.savioli,miguel.silvavieira,pablo.lamata\}@kcl.ac.uk} \\
 \IEEEauthorrefmark{2}\texttt{g.montana@warwick.ac.uk} }

\maketitle

\begin{abstract}

The segmentation of the left ventricle (LV) from CINE MRI images is essential to infer important clinical parameters. Typically, machine learning algorithms for automated LV segmentation use annotated contours from only two cardiac phases, diastole, and systole. In this work, we present an analysis work-flow for fully-automated LV segmentation that learns from images acquired through the cardiac cycle. The workflow consists of three components: first, for each image in the sequence, we perform an automated localization and subsequent cropping of the bounding box containing the cardiac silhouette. Second, we identify the LV contours using a Temporal Fully Convolutional Neural Network (T-FCNN), which extends Fully Convolutional Neural Networks (FCNN) through a recurrent mechanism enforcing temporal coherence across consecutive frames.
Finally, we further defined the boundaries using either one of two components: fully-connected Conditional Random Fields (CRFs) with Gaussian edge potentials and Semantic Flow. Our initial experiments suggest that significant improvement in performance can potentially be achieved by using a recurrent neural network component that explicitly learns cardiac motion patterns whilst performing LV segmentation.
\end{abstract}

\section{Introduction}

Cardiovascular disease (CV) remains the leading cause of death worldwide \cite{KreatsoulasC}.
Cardiac Magnetic Resonance (CMR) is frequently used for diagnostics of various cardiac diseases \cite{SalernoM}. Segmentation of the Left Ventricle (LV) from CMR images provides a standard procedure for the determination of cardiac parameters which is time-consuming and requires relevant experience.

The segmentation of the heart requires the identification of two anatomical regions: the inner part called the endocardium, and the outer part, the epicardium. Identifying and segmenting those two regions in CMR images presents different levels of difficulty: while the endocardium has sufficient contrast, the epicardial surface presents profiles of intensity with little contrast \cite{ValliappanRaman}.

Notwithstanding, a clear delineation of the epicardium contours is a challenging task due to non-homogeneity in the blood flow. Moreover, the presence of strong unevenness on the wall in the interior of the heart chambers due to the sporadic presence of papillary muscles does not allow a clear delimitation of the endocardial wall \cite{EbelingC}. 
The level of difficulty also depends on the particular ventricular section; the apical and basal sections are much more difficult to segment because of the resolution of the image is lower and does not allow the detection of ventricular structure \cite{CarolinePetitjean}. 

Fully automatic segmentation algorithms have been proposed in the last decade to speed up segmentation, but the robust and reliable solution is still missing for clinical practice \cite{Olivier}. 
Existing approaches for fully-automated LV segmentation can be divided into three groups: image-based methods, deformable models, and pixel-based classification methods. Image-based methods consist of finding the endocardium border using a threshold of gray level intensities \cite{AGoshtasby} or Dynamic Programming (DP) \cite{JYYeh}. The Level-Set (LS) segmentation, such as \cite{TFChan}, is the most successful choice within the methods using deformable models. 

Finally, pixel classification methods, based on supervised Deep Neural Networks \cite{Ngo1, Avendi, Tran2016AFC, RudraP}, have become relatively successful for both object detection and segmentation. Proposed architectures include Fully Convolution Neural Networks (FCNN) \cite{Tran2016AFC}, Stacked Autoencoders (SA) \cite{Avendi} and Deep Belief Networks \cite{Ngo1}. 

Occasionally, a post-processing method is also applied, following the initial segmentation, to obtain more robust and accurate contours. In some cases, a single architecture trained end-to-end has been shown to achieve satisfactory performance without the need for further pre- and post-processing, e.g. by modeling the spatial dependence amongst SA sections \cite{RudraP}. There are also architectures for segmentation of three-dimensional images, such as \cite{JianxuChen}, which have shown to boost the performance of 2D approaches \cite{Olivier}.

 Also, Optical Flow (OP), i.e. the pattern of apparent motion of objects, has also been used in the segmentation context to exploit the regularity in the cardiac motion and constrain the degree of allowed LV movement between adjacent image frames. Indeed, OP equations have also been combined with the LS equations to enforce visual constancy \cite{Paragios}.

In this work, we exploit a complementary source of information, the coherence across consecutive frames, and propose a Temporal Convolution Neural Network (T-FCNN) architecture with Semantic Flow post-processing. The performance of the proposed architecture is compared against an established FCNN model, which treats each cardiac phase independently and achieves good segmentation performance \cite{Tran2016AFC}. An alternative post-processing choice, the use of Conditional Random Fields (CRFs)  \cite{NIPS2011_4296}, is also investigated.

\section{TWINS-UK dataset} 

The TWINS-UK is a voluntary registry that includes \textgreater 12,000 twins \cite{MoayyeriA}. For this study, 68 consecutive female subjects (mean age $62 \pm 9$ years) were recruited from the TWINS-UK cohort. 

The CMR scans were performed on a 1.5-T clinical scanner (Achieva, Philips Healthcare, Best, The Netherlands). Each dataset included 12 to 14 equidistant and contiguous short-axis CINE from the atrioventricular (AV) ring to the apex, completely covering both ventricles (slice thickness 8 mm; no gap mm; field of view was $360 \times 480$ mm and matrix size $156 \times 144$).

The ECG-gated steady-state free-precession (SSFP) end-expiratory breath-hold 2D CINES were acquired. Images were acquired with 30 phases/cardiac cycle corresponding to a temporal resolution of 25-35 milliseconds at a heart rate of 60-80 beats per minute.

The dataset was randomly divided into training, validation and testing sets of sizes $70\%$, $15\%$ and $15\%$, respectively. Each pair of twins was allocated to a specific subset and not separated to avoid any genetic similarities affecting our results.

\section{Proposed analysis work-flow}

The work-flow is divided into three stages: LV position detection, LV segmentation, and LV contour refinement. Our input $x_{t,i}$ consists of the entire temporal sequence obtained in all cardiac phases, while the label output $y_{t,i} $ is the corresponding sequence of binary masks, one per time point. Whereas, the $i$ index indicate the specific label pixel at the temporal image phase $t$.  
Each input image was downsized to $236 \times 236$. The output masks have the same size after a padding operation of $44 \times 44$ pixels. 

\subsection{Single-frame LV position detection}

For the automated detection of the LV in each time frame, we used the Over-Feat algorithm \cite{PierreSermanet}. 
In order to train the object detection layers in Over-Feat, we pre-trained the GoogLeNet architecture \cite{ChristianSzegedy} within the ImageNet database and later carried out fine-tuning using our LV images in a sliding window fashion. The prediction of the bounding box coordinates is obtained through regression layers minimizing an L2 loss. 

\subsection{Sequence-based LV segmentation}

Upon detecting the bounding box containing the heart, the cardiac contours are inferred with an architecture that extends U-Net \cite{Ronneberger2015}, originally proposed for the segmentation of biomedical images. 

A standard U-Net takes an individual frame as input and estimates the corresponding binary mask as output. An encoding (descending) path is composed of a sequence of hidden layers that are used to learn a representation of the input image. Each hidden layer is represented by repeated $3\times3$ convolution operation following by pooling and Rectified Linear Unit (ReLU).

More specifically, our solution is an improvement of a Fully Convolutional Neural Network (FCNN) \cite{ShelhamerLong}, which becomes our baseline for comparisons. 
This is then followed by another sequence of hidden layers forming a decoding (ascending) path through which all the feature maps are gradually restored to the original image size and are used to infer the final binary masks. Similar to U-Net, the hidden layers in the encoding path, and the decoding path consisting of the convolutional (or upsampling) filters also followed by pooling and ReLU mappings.
 
One of the key features of this architecture is the use of skip paths connections between convolution and deconvolution layers, for the purpose of fusing global and local information. Every skip paths concatenate feature maps from the encoding to the decoding path (i.e extracting feature maps from each convolutional encoding block and concatenating with those of the decoding deconvolutional block). 

\begin{figure*}[ht]
  \centering
    \includegraphics[width=7in]{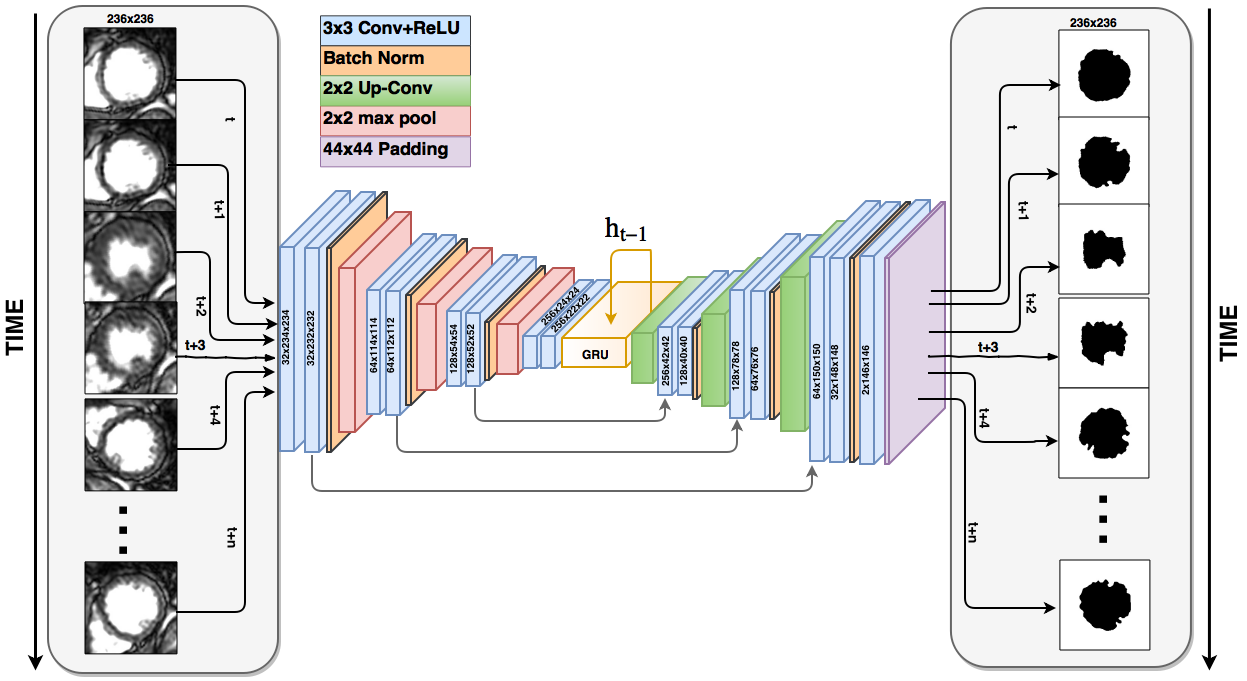}
      \caption{T-FCNN architecture. Blue blocks represent a convolutional layer followed by ReLU operations; orange blocks represent batch normalization operations; green blocks correspond to up-convolution operations; pink blocks max-pooling operations and purple blocks identify padding operations. Black arrows represent the input and output for each CINE MRI frame and its temporal segmentation. Gray arrows denote copy operations and the yellow arrow indicates the recurrent connection implemented through a Conv-GRU layer.}
  \label{fig:workflowedge}
\end{figure*}

For the purpose of segmenting the entire cardiac motion, we modify the U-Net architecture by adding a recurrent layer immediately after the descending path. The purpose of this layer is to leverage the information flowing from preceding image frames and enforce temporal coherence (i.e the temporal redundancy over time). Then, the feature maps learned at the encoding stage are used as inputs for a recurrent element coded with a Convolutional Gated Recurrent Unit (Conv-GRU) \cite{JunyoungChung}. The resulting architecture, a Temporal Fully-Convolutional Neural Network (T-FCNN), is illustrated in Fig. \ref{fig:workflowedge}. 

The encoding path consists of four blocks of hidden  layers, which are arranged as follows: two $3 \times 3$ convolutional layers (with stride set to $1$), a ReLU layer, a BN layer and, finally, a $2 \times 2$ max pooling layer (with stride set to $2$).

\subsection{Post-processing}

The final component of the segmentation work-flow is a post-processing algorithm that has the potential to further improve upon the binary masks predicted by the T-FCNN. In this study, we compare two methods that have been particularly successful for semantic segmentation tasks, i.e. Fully Connected CRFs with Gaussian edge potentials \cite{NIPS2011_4296} and Semantic-Flow (SF) \cite{SevillaLaraSJB16}.
The fully connected CRFs minimise an energy function: 
$$
\text{E}(x)= \sum_{t=0}^{T-1} (\sum_{i} \theta_{i} (y_{t,i}) + \sum_{ij} \theta_{ij} (y_{t,i},y_{t,j}))
$$
Where $y_{t,i}$ is the segmentation mask from frame $t$, $i$ is the index of each label pixel, $\theta_{i}$ is the unary potential, defined as $\theta_{i}(y_{t,i})=-\log( \text{P}(y_{t,i}))$, and $\theta_{ij}$ is the pairwise potential, defined by:  
$$
 \theta_{ij} (y_{t,i},y_{t,j}) = \mu (y_{t,i},y_{t,j}) k (f_{t,i},f_{t,j}).
$$
The function $\mu (y_{t,i},y_{t,j})$ equals one when $y_{t,i} \neq y_{t,j}$, zero otherwise, and the function $k (f_{t,i},f_{t,j})$ is a Gaussian kernel, evaluated using features $f_{t,i}$ and $f_{t,j}$ corresponding to pixels $i$ and $j$ respectively.
Notably, the first kernel is driven by both pixels position ($p$) and color intensities ($I$); where the second kernel only uses pixels position. 
Kernel coefficients were kept fixed, and the energy function was minimized using the L-BFGS algorithm for nonlinear optimization using multi-thread CPUs. 
The second post-processing algorithm uses a Semantic-Flow (SF) approach, which is another alternative to exploit the temporal coherence in a sequence. Our implementation follows the formulation presented in \cite{SevillaLaraSJB16}, and the reader is referred to that paper for a detailed explanation of the concepts that are summarised next.
Given a CINE MRI sequence of 2D frames $x_{t,k}$, our aim is to estimate simultaneously the flow vector field $(u_{t,k},v_{t,k})$ that maps every pixel between consecutive 2D images. The task is divided in two regions defined by $g_{t,k}$ for $k \in \{1,2\}$, corresponding to LV mask and background, and vector fields are parametrised with a set of parameters $\theta_{t,k}$. 
Input is $y_{t,k}$, the initial LV segmentation comes form T-FCNN. We then wish to minimise the following energy function $E_{LV_{Seg}}(u,v,g,\theta,x,y)$, that consists of five terms: data, motion, time, space and coupling term.

\begin{equation*}
\begin{split}
E_{LV_{Seg}}(u,v,g,\theta,x,y) =  \sum_{k=0}^{2} \Bigg\{  \sum_{t=0}^{T-1} \Big\{ E_{d}(u,v,g,x_{t},x_{t+1}) 
\\ + \lambda_{m} E_{m}(u,v,g,\theta) 
+ \lambda_{t} E_{t}(u,v,g_{t},g_{t+1})\Big\} + \sum_{t=0}^{T} \\ \Big\{ \lambda_{c} E_{c}(g,y)  + \lambda_{s} E_{s}(g) \Big\}\Bigg\}
\end{split}
\end{equation*}

The data term $E_{d}$ measures the similarity between the gray scale intensities of adjacent frames, the motion term $E_{m}$ enforces some regularity of the vector field in pixels that belong to the same region or that are close to each other, the time term $E_{t}$ constrains the regularity of pixels belonging to a LV region or background along the sequence, $E_{s}$ enforces the connectivity of pixels within the same region, and the coupling term $E_{c}$ emphasizes the affinity between background segmentation and LV segmentation. The different $\lambda$ are constants that weight the contribution of the energy terms, and that are left constant in our study. 

\begin{figure*}[ht]
  \centering
    \includegraphics[width=1\textwidth]{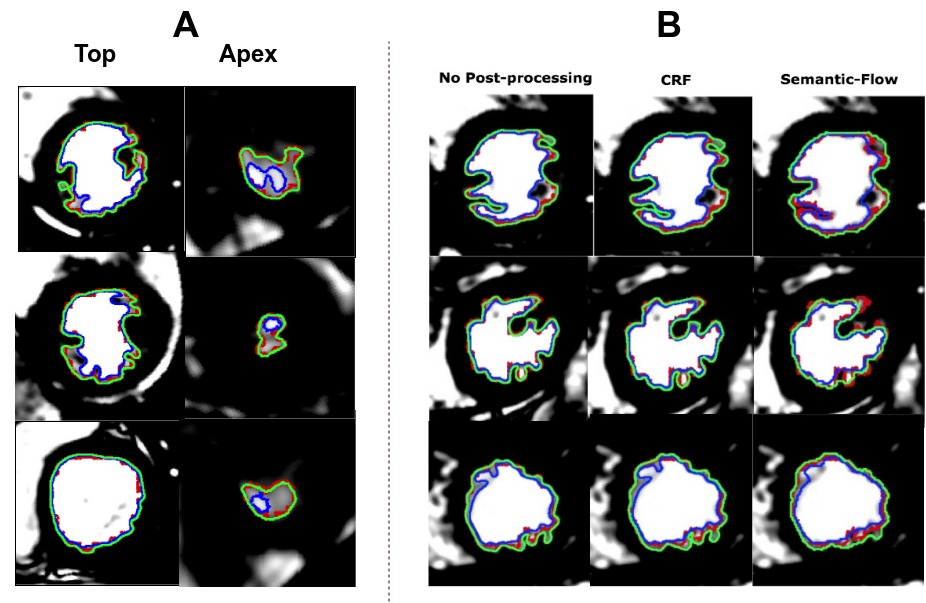}
   \caption{a) Comparison of the segmentations obtained from FCNN (blue line) vs T-FCNN (green line) compared within clinical ground truth (red line). The left column shows the top slices LV segmentation. While the right column shows the apex LV segmentation cases. Both, show that T-FCNN has good segmentation performance in comparison with FCNN. Especially FCNN, it tends to segment only high-intensity regions; as we can see in the apical cases.
b) Comparison of the segmentation obtained within no post-processing methods vs post-processing. As we can see the Semantic-Flow (Semantic flow column) tends to blunt the LV prediction, while the CRF (CRF column) remains fairly consistent with T-FCNN (no post-processing column).}
  \label{fig:results_APD}
\end{figure*}

\section{Experimental Results}

The performance of the detection of the position of the LV was assessed with the Intersection Over Union, reaching a score of 98\%. The accuracy of the final segmentation was measured using three different metrics: the Dice Index (DI), the Average Perpendicular Distance (APD) and the Conformity (C) index \cite{Avendi}. 

A set of architectures from the baseline FCNN to the proposed T-FCNN with CRFs was implemented and analyzed. 
Parameters of both T-FCNN and FCNN have tuned within adadelta optimizer, and both architectures are trained end-to-end with spatial cross entropy criterion with a learning rate of $1e-4$.
 
Results are reported in [Table \ref{table:tabresults}]. T-FCNN improves FCNN, reducing a 30.5\% the APD metric (from $10.3$ to $7.1$ mm), and the addition of CRFs further reduces another 12\% this error metric (from $7.1$ to $6.3$ mm). The CRF did improve the performance in all cases, but SF only improved the metric of APD when added to the FCNN. Some illustrative examples of the final segmentation result are provided in [Fig. \ref{fig:results_APD}]. 

\begin{table*}[ht]
\small
  \begin{center}
   \begin{tabular}{|c|c|c|c|}
  \hline 
   Algorithms & DICE (\%)  & APD (mm) & C(\%) \\  
  \hline 
  FCNN & 0.9745(0.0163) & 10.2734(12.7622) & 0.9472(0.0352) \\  
  T-FCNN & 0.9803(0.0263) &  7.1427(11.0284) & 0.9583(0.0592) \\  
  FCNN+CRFs & 0.9774(0.0161) & 8.2084 (8.3545) & 0.9532(0.0345) \\ 
  \textbf{T-FCNN+CRFs} & \textbf{0.9815(0.0245)} & \textbf{6.2903(8.3814)} & \textbf{0.9610 (0.0551)} \\ 
  FCNN+SF & 0.9735(0.0506)       & 9.3289(13.4287)    & 0.8872(1.2804) \\ 
  T-FCNN+SF & 0.9717(0.0743)     & 8.1635(12.8057)    & 0.8213(1.7954)  \\ 
  FCNN+CRFs+SF & 0.9762(0.0462)   & 8.4325(11.1094)    & 0.8939(1.2843) \\ 
  T-FCNN+CRFs+SF & 0.9737(0.0735) & 7.5983(11.5803)    & 0.8097(1.9614) \\ 
  \hline 
  \end{tabular}
  \vspace{2mm}
  \caption {Segmentation performance results obtained on the TwinsUK dataset using different combinations of segmentation (FCNN and T-FCNN) and post-processing (CRF and SF) algorithms.}\label{table:tabresults}
\end{center}
\end{table*}

\section{Discussion and Conclusions}

Exploitation of the temporal coherence across frames is a useful resource for the fully automated, robust and accurate segmentation of CINE MRI sequences, as required for clinical practice. 

A CINE MRI study has a large level of coherence both in space and time. Driven by clinical interest (i.e. the generation of metrics such as the blood pool volume at a given time) and by the less burden required for the generation of the ground truth (i.e. only need to segment 10-12 slices at a given time), most of the algorithms proposed up to date exploit the coherence in space \cite{RudraP}. 

The use of a recurrent unit in space (i.e. across slices) did reduce the APD in a 4.2\% or a 13\% at the MICCAI and PRETERM cohorts as reported in \cite{RudraP}, whereas the use of a recurrent unit in time (i.e. across frames) improved the APD in a 30\% in our TWINS-UK cohort (see Table \ref{table:tabresults}). Temporal coherence seems thus to be a more useful resource, probably due to the fact that there is indeed a larger correlation between adjacent frames in time than between adjacent slices in a CINE MRI study. And this has been achieved by a simple extension of the U-Net architecture through a recurrent neural network component (T-FCNN). 

In the search for the best strategy to exploit temporal coherence, our results suggest that T-FCNN is a better solution than an FCNN+SF. We can't claim that we fully exploited the potential of both strategies (i.e. exploration of all the parametric space in SF is a tedious exercise), but our results with a reasonable set of parameter and architecture combinations consistently suggested that SF did at most only marginally improve the LV segmentation performance. 

It is important to note that the clinical protocol for the CMR segmentation in out TWINS-UK cohort included the papillary muscles as part of the myocardium, in contrast with the majority of previous studies \cite{RudraP, Olivier}. This is a more challenging task for the human observer and the algorithm, since the smoothness constraints of contouring a circular shape cannot be used, and is the reason why previous APD values were smaller (i.e. around 2mm \cite{RudraP}). 

Nevertheless, the use of a recurrent unit is not a perfect solution, since it is going to be limited by the problem of the vanishing gradients that reduces the performance of the backpropagation through time. This is going to limit the number of cardiac frames that can be imputed to the architecture. In practice, this should not be a major problem, since acquisitions with more than 30 frames, as used in this study, are not common in CMR. 

The translation of this concept to echocardiography will nevertheless bring some challenges, where you can deal with sequences of up to thousands of frames using ultra-fast acquisition protocols. An easy first solution will then be the use of an upper bound value to the gradient (i.e. gradient clipping) or adding a regularisation term able to increase or decreases the gradients magnitude \cite{Razvan}. Finally, the use of 3D convolution kernels could be useful for progressively decrease the number of input sequences; avoiding excessively long temporal or spatial sequences.  

The main practical bottleneck is the heavier burden of segmentation needed to transfer the learning to another study or experimental setting. Instead of learning from single frames, the proposed architecture will need to learn from complete sequences. The availability of semi-automatic segmentation solutions in commercial products should make this task affordable. In any case, our TWINS-UK dataset with its manual ground truth segmentation is made available under request to the community as a reference for future solutions to exploit the temporal coherence in CMR sequences.

\end{document}